\def\year{2022}\relax
\documentclass[letterpaper]{article} 
\usepackage{array}
\usepackage{aaai22}  
\usepackage{times}  
\usepackage{helvet}  
\usepackage{courier}  
\usepackage[hyphens]{url}  
\usepackage{graphicx} 
\usepackage{natbib}  
\usepackage{caption} 
\DeclareCaptionStyle{ruled}{labelfont=normalfont,labelsep=colon,strut=off} 
\usepackage{multirow}
\usepackage{bm}
\usepackage{algorithm}
\usepackage{algorithmic}

\usepackage{newfloat}
\usepackage{listings}
\floatstyle{ruled}
\newfloat{listing}{tb}{lst}{}
\floatname{listing}{Listing}

\usepackage{subfigure}
\usepackage{color}
\usepackage{amssymb}
\usepackage{multirow}
\usepackage{amsmath}

\title{UFPMP-Det: Toward Accurate and Efficient Object Detection on Drone Imagery}
\author {
    Yecheng Huang\textsuperscript{\rm 1, \rm 2},
    Jiaxin Chen\textsuperscript{\rm 2},
    Di Huang\textsuperscript{\rm 1, \rm 2}\thanks{~indicates the corresponding author.}
}
\affiliations {
    \textsuperscript{\rm 1} State Key Laboratory of Software Development Environment, Beihang University, Beijing, China\\
    \textsuperscript{\rm 2} School of Computer Science and Engineering, Beihang University, Beijing, China\\
    \{ychuang, jiaxinchen, dhuang\}@buaa.edu.cn
}
\usepackage{bibentry}

\begin{document}

\maketitle

\begin{abstract}

This paper proposes a novel approach to object detection on drone imagery, namely Multi-Proxy Detection Network with Unified Foreground Packing (UFPMP-Det). To deal with the numerous instances of very small scales, different from the common solution that divides the high-resolution input image into quite a number of chips with low foreground ratios to perform detection on them each, the Unified Foreground Packing (UFP) module is designed, where the sub-regions given by a coarse detector are initially merged through clustering to suppress background and the resulting ones are subsequently packed into a mosaic for a single inference, thus significantly reducing overall time cost. Furthermore, to address the more serious confusion between inter-class similarities and intra-class variations of instances, which deteriorates 
detection performance but is rarely discussed, the Multi-Proxy Detection Network (MP-Det) is presented to model object distributions in a fine-grained manner by employing multiple proxy learning, and the proxies are enforced to be diverse by minimizing a Bag-of-Instance-Words (BoIW) guided optimal transport loss. By such means, UFPMP-Det largely promotes both the detection accuracy and efficiency. Extensive experiments are carried out on the widely used VisDrone and UAVDT datasets, and UFPMP-Det reports new state-of-the-art scores at a much higher speed, highlighting its advantages.

\end{abstract}

\section{Introduction}
\begin{figure}[!ht]
    \centering
    \subfigure[Instance scales in MS COCO (left) \emph{vs.} VisDrone (right)]{
    \includegraphics[width=0.95\linewidth]{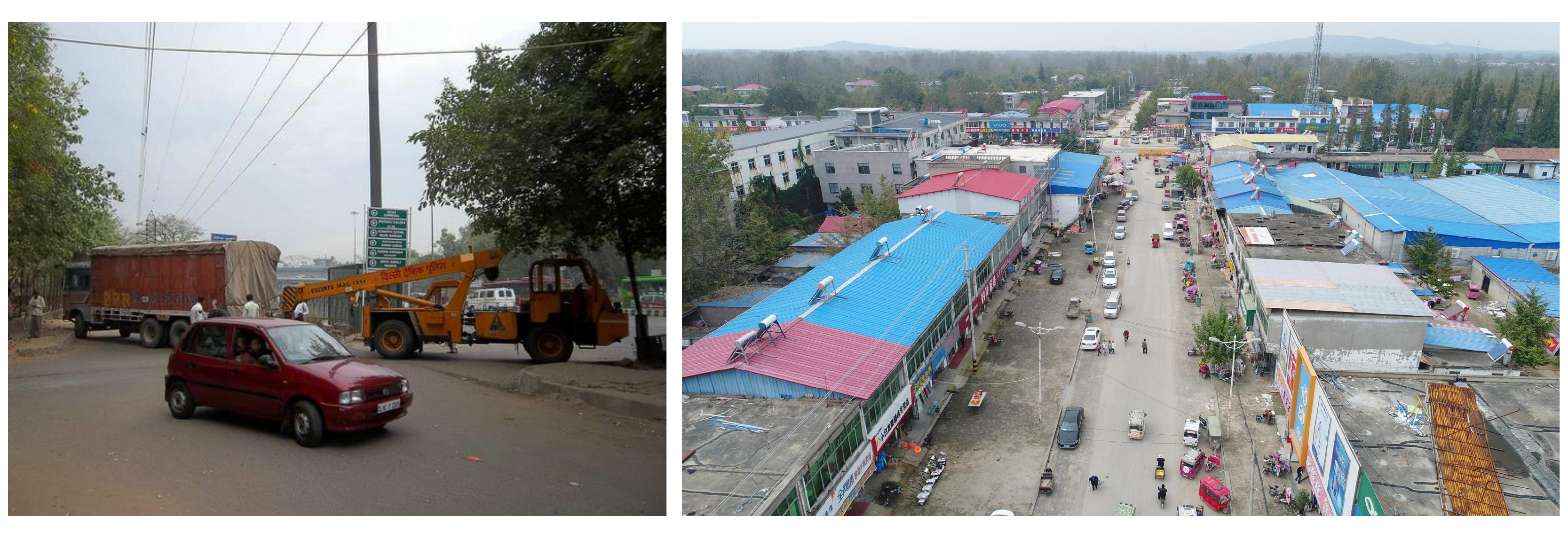}
    }
    \subfigure[Semantically similar categories on VisDrone]{
    \includegraphics[width=0.95\linewidth]{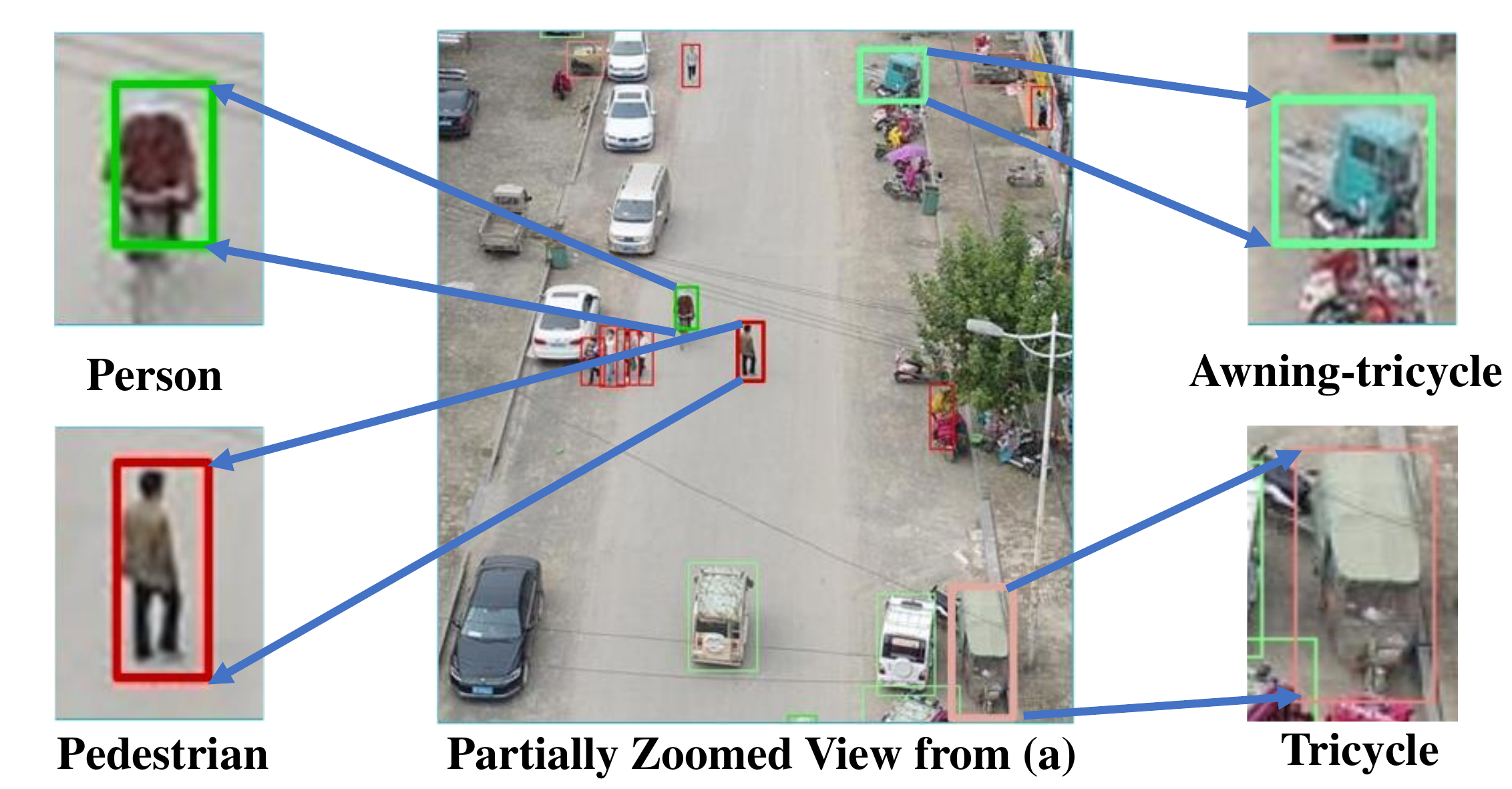}
    }
    
    \caption{Challenges in object detection on drone imagery.}
    \label{pie of ratio}
    \vspace{-3mm}
\end{figure}
Recently, the drone, also known as UAV, has become a popular equipment for its trade-off between mobility and stability in many applications, such as security surveillance, aerial photography, express delivery, and agricultural production, where drone image based object detection is the fundamental issue. Although object detection on natural images has been greatly developed by Convolutional Neural Networks (CNNs) during the past several years, that on drone imagery is still limited in terms of both accuracy and efficiency.  

One major challenge to detect objects using drone images lies in that there exist \textbf{a large amount of instances of very small sizes}, and compared with the case in the benchmarks of general object detection, \emph{e.g.}, PASCAL VOC \cite{VOC} and MS COCO \cite{COCO}, the ratio of such instances is rather high, as Fig.1 (a) depicts. Meanwhile, computing resources are usually inadequate in drones, current time-consuming solutions, like image pyramid, which proves effective in general object detection, are no longer competent. Instead, a coarse-to-fine pipeline is often followed \cite{DMNet, ClutDet}, where a coarse detector is launched to locate the large-scale instances and sub-regions that contain densely distributed small ones and a fine detector is further applied to those regions to find instances of small sizes. These methods show promising results; however the sub-regions delivered by the coarse detectors are relatively rough, with a large portion of backgrounds included, incurring inefficient computations. Furthermore, since they partition the input image into multiple chips, they have to individually process each sub-region, leading to several times of inferences for final decision. The two drawbacks hence suggest room for efficiency improvement.

Another considerable challenge is that \textbf{some object categories defined in drone datasets}, \emph{e.g.} VisDrone \cite{VisDrone}, \textbf{are semantically close to each other}, for instance, pedestrian \emph{vs.} person; tricycle \emph{vs.} awning-tricycle, and the appearances of instances belonging to these categories are quite confusing, as displayed in Fig. 1 (b). Besides, due to more severe disturbances caused by flying altitude, viewing angle, and weather condition, the distances between features of the instances from the same category tend to be enlarged. The inter-class similarities and intra-class variations are thus more seriously intertwined than in general object detection, making the classification of instances even harder. Unfortunately, to the best of our knowledge, this problem is ignored in the previous literature, which leaves much space for accuracy amelioration.

 To address the two challenges aforementioned, \emph{i.e.} high percentage of small instances and low distinctiveness of similar categories, in this paper, we present a novel approach to object detection on drone images, namely Multi-Proxy Detection Network with Unified Foreground Packing (UFPMP-Det). It substantially extends the coarse-to-fine framework by two specially designed modules. For the former, the \textbf{Unified Foreground Packing (UFP)} module is proposed. UFP operates in a two-stage manner, where the foreground sub-regions by the coarse detector are firstly merged through a clustering algorithm to suppress backgrounds and the resulting regions are then packed into a mosaic with adaptively enlarged scales. By this mean, the foreground ratio of small objects are increased, and the successive fine detector performs inference only once at the mean time. As a consequence, both the detection accuracy and speed can be significantly promoted. For the latter, the \textbf{Multi-Proxy Detection Network (MP-Det)} module is proposed. In MP-Det, the multi-proxy learning scheme originally explored for the image retrieval task is adapted to object detection, aiming to boost the performance of the classification head by generating compound and flexible decision boundaries. In particular, to bypass the collapse phenomenon in training multiple proxies, Bag-of-Instance-Words (BoIW) guided Optimal Transport is introduced, in which BoIW well models the distribution of each category in the presence of the confusion between inter-class similarities and intra-class variations, thus facilitating feature-proxy matching by Sinkhorn optimization. We extensively evaluate the proposed approach on two public databases, \emph{i.e.} VisDrone and UAVDT, and report the state-of-art scores with largely promoted efficiency, highlighting its effectiveness.

\section{Related Work}
\subsection{Generic Object Detection} 
Generic Object Detection has been largely developed in recent years along with the success of CNNs within the community of artificial intelligence, especially computer vision. On whether using pre-defined sliding windows or proposals, the existing methods are divided into two main streams, \emph{i.e.} anchor-based and anchor-free. The \textbf{anchor-based} detectors sample the box space into discrete bins and refine the boxes of objects accordingly, and anchors are taken as regression references and classification candidates to infer proposals in multi-stage detectors, such as R-CNN \cite{R-CNN}, Faster-RCNN \cite{faster-rcnn} and Cascade-RCNN \cite{cascade-rcnn}, or final bounding boxes in single-stage ones, \emph{e.g.} SSD \cite{SSD}, YOLO \cite{YOLOv3}, and RetinaNet \cite{RetinaNet}. Compared to anchor-based ones, the \textbf{anchor-free} detectors avoid complicated computation related to anchor boxes and bypass the corresponding prior hyper-parameter setting, figuring out a promising alternative, where FCOS \cite{FCOS}, FSAF \cite{FSAF}, and GFL v1/v2 \cite{GFLV1, GFLV2} are representatives.

\subsection{Object Detection on Drone Imagery}
Despite the progress achieved by object detection on natural images, \emph{e.g.} PASCAL VOC and COCO, the performance on drone images is still far from satisfactory. As stated, the high percentage of small instances and low distinctiveness of similar categories make the issue more challenging. Inspired by the region search strategies \cite{Sniper, autofocus} employed in general object detection to accelerate training and inference, all the current studies focus on the problem of small instances and address it by following a coarse-to-fine framework, which serially adopts a simple strategy or a coarse detector to roughly suggest the regions with densely distributed small instances and a fine detector to precisely localize the objects on them. For example, in \cite{tiling}, the tiling based method makes even splits to produce sub-regions but it tends to break instances while truncating images; ClutDet \cite{ClutDet} applies a sub-network to crop sub-regions from the raw input; and DMNet \cite{DMNet} estimates the object density in the original image and then separates sub-regions as minimal areas of connected possible blocks.

Such methods indeed advance drone image based object detection; however, the sub-region generated are not so decent with much background in them and the decision on an entire image requires multiple inferences on its sub-regions, both of which show room for efficiency. On the other side, the low distinctiveness of similar categories has not been discussed, resulting in limitation to improved accuracies.

\subsection{Small Object Detection}

Since MS COCO \cite{COCO} was released, small object detection has become a critical topic and received increasing attention. FPN \cite{FPN} is a major choice to handle scale changes through feature pyramid, and it is extended to a number of variants, including EFF-FPN \cite{EFF-FPN}, AugFPN \cite{AugFPN}, \emph{etc.} Perceptual GAN \cite{PGAN} utilizes an adversarial network to boost the detection performance by narrowing the representation difference between small and large objects, and a super-resolution feature generator is trained with proper high-resolution target features for supervision. A similar idea appears in \cite{better2follow} but considers the impacts of the receptive fields of various sizes as well. TinyPerson \cite{tinyperson} claims that scale mismatch between the data for network pre-training and detector learning incurs degradation and designs Scale Match to align object scales between different datasets.

The methods above are validated either on MS COCO for general object detection or on other detection tasks for special objects, \emph{i.e.} persons or traffic signs. As we analyze before, drone image based object detection has its unique challenges, where the scale distribution of instances on drone images is quite different and the object categories share semantic similarity as shown in Fig. 1 (a) and (b), thereby making them problematic to the given issue.


\begin{figure}[t]
    \centering
    \subfigure[Image Pyramid]{\includegraphics[width=1\columnwidth]{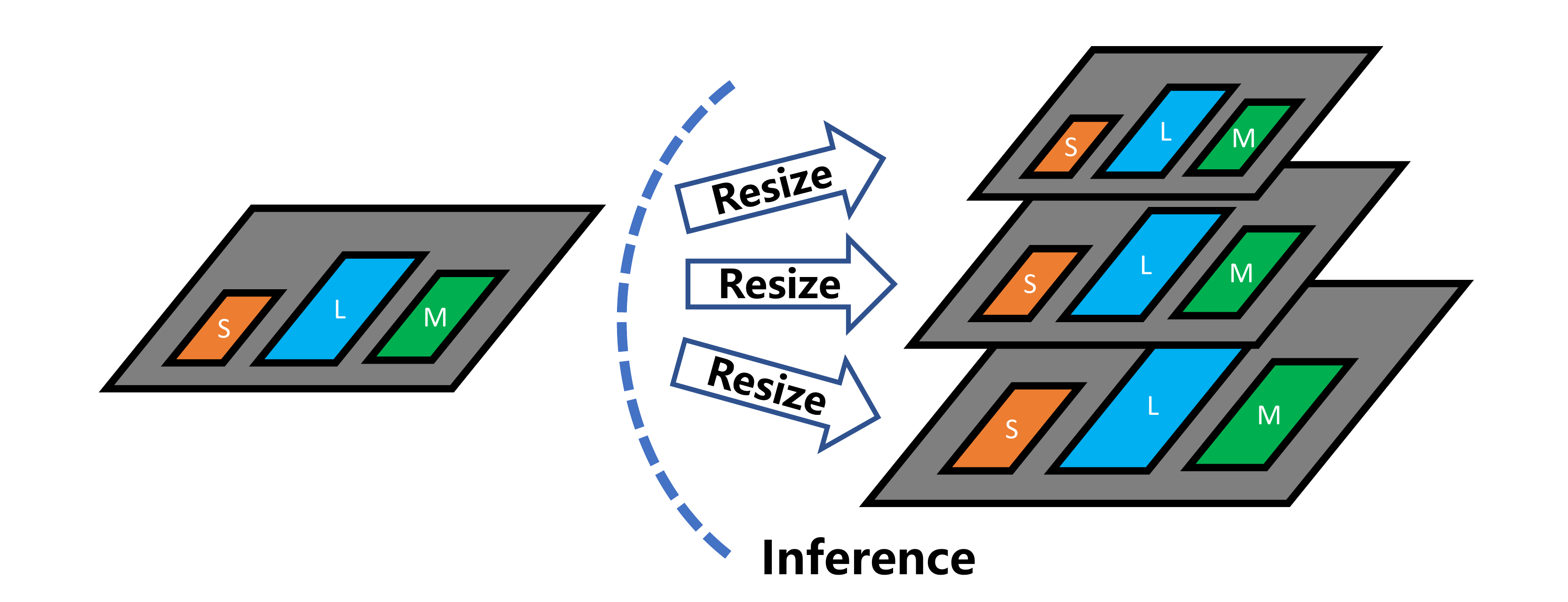}}
    \subfigure[Cluster Region]{\includegraphics[width=1\columnwidth]{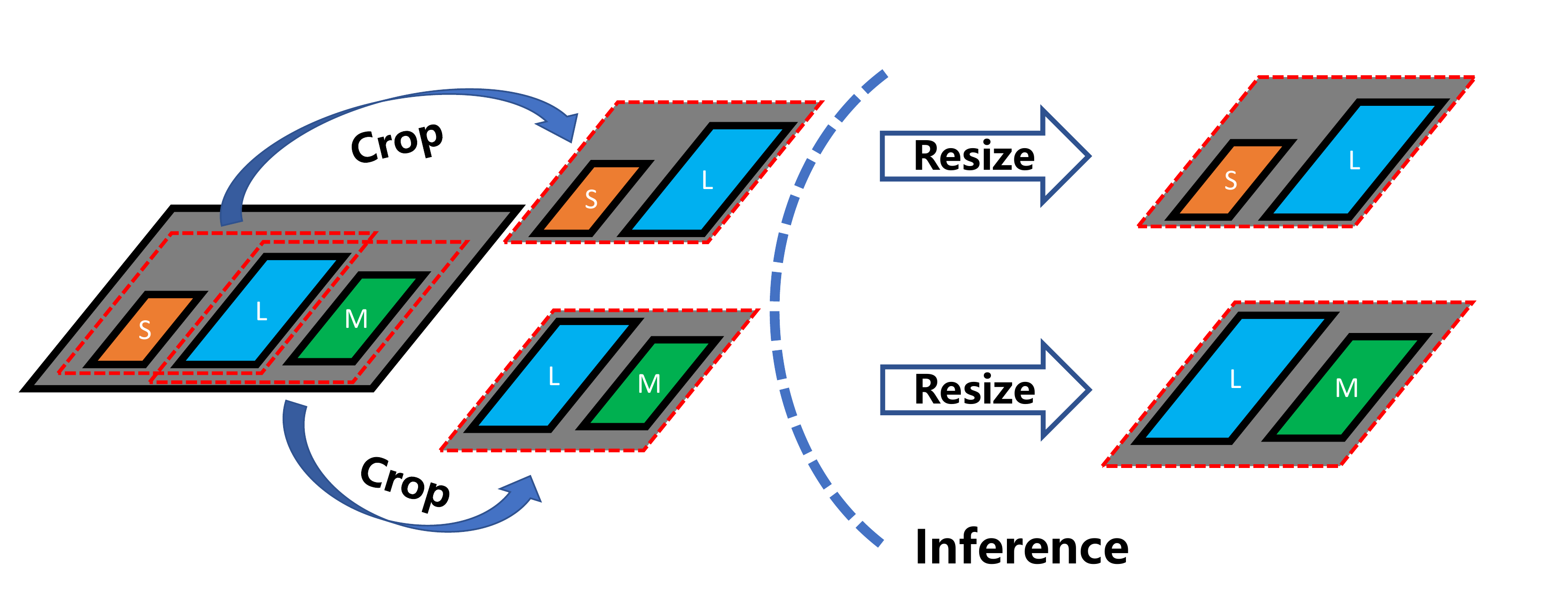}}
    \subfigure[Our UFP]{\includegraphics[width=1\columnwidth]{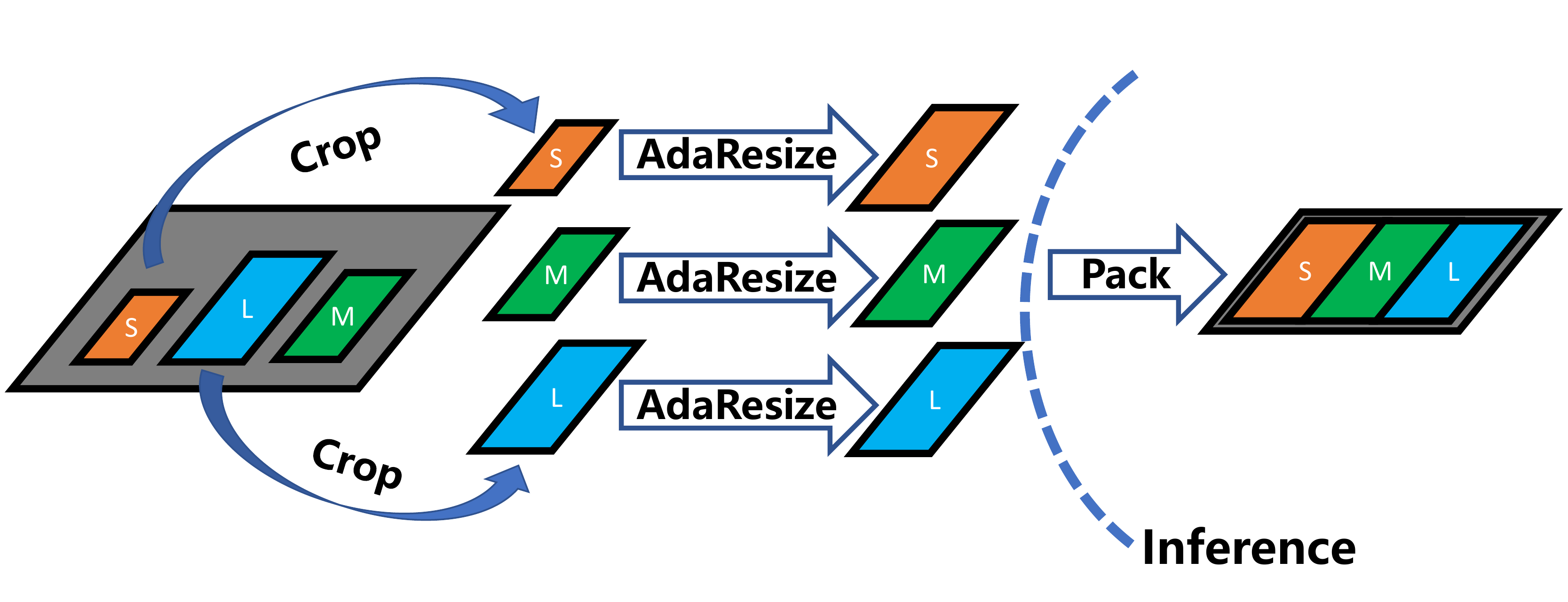}}
    \caption{Comparison of different pipelines when handling scale changes. Orange, green and blue boxes indicate small, median and large scale objects, respectively.}
    \label{comparison with pipeline}
\end{figure}
\vspace{-3mm}
\section{The Proposed UFPMP-Det Approach}

To tackle the two challenges, \emph{i.e.} high percentage of small instances and low distinctiveness of similar categories, a novel approach, namely Multi-Proxy Detection Network with Unified Foreground  Packing (UFPMP-Det), is proposed. It contains two major stages, where in the first stage, the Unified Foreground Packing (UFP) module converts raw drone images into mosaics with higher foreground ratios, and in the second stage, the Multi-Proxy Detection Network (MP-Det) module infers on the mosaic images, which employs a multi-proxy learning scheme with Bag-of-Instance-Words (BoIW) guided Optimal Transport to model complex object distributions. They are introduced in detail in the following.




\subsection{Unified Foreground Packing}

\begin{figure}[!t]
    \centering
    \includegraphics[width=1\columnwidth]{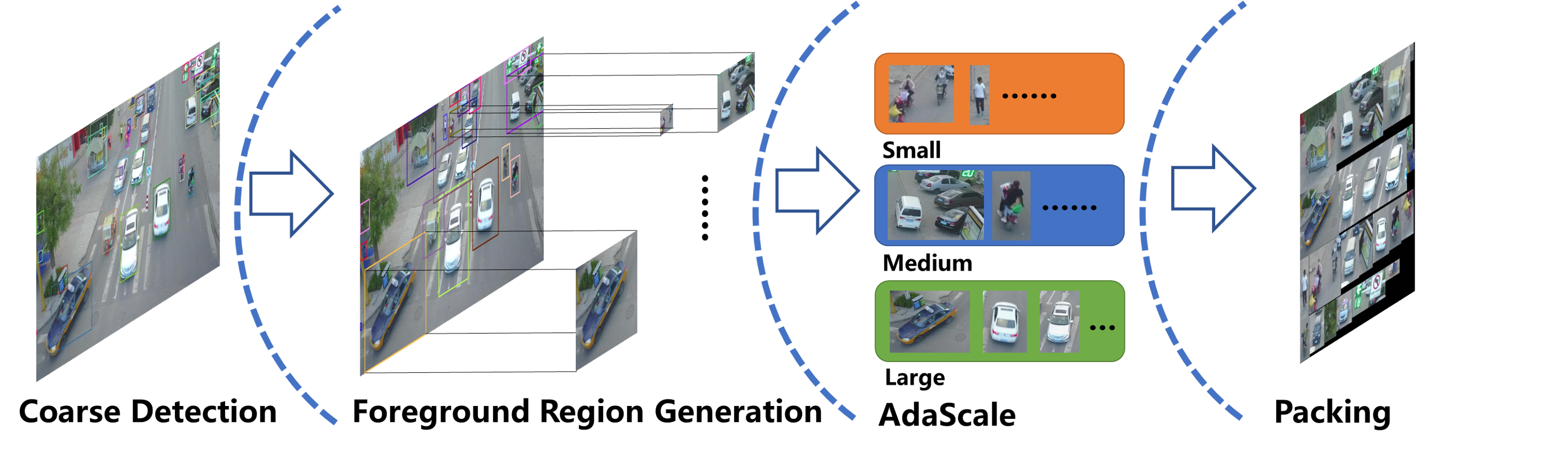}
    \caption{Pipeline of Unified Foreground Packing (UFP).}
    \label{ufp pipeline}
    \vspace{-3mm}
\end{figure}

The UFP module aims to convert original drone images into unified mosaic ones with significantly increased foreground ratios and enlarged sizes of small objects. As Fig.~\ref{ufp pipeline} shows, when foreground sub-areas are extracted from the drone image by a coarse detector, UFP introduces three successive steps: (1) foreground sub-areas are merged to several clustered ones; (2) small scale cluster regions are enlarged adaptively; and (3) adjusted cluster regions are packed into a unified mosaic.


\subsubsection{Foreground Region Generation}

\begin{algorithm}[!t]
\caption{ Foreground Region Generation}
\label{Cluster Region Generating algorithm}
\textbf{Input}: Bounding boxes $\mathcal{B}_{c} $\\
\textbf{Output}: Merged regions $\mathcal{B}_{r}$
\begin{algorithmic}[1] 
\STATE Initialize $\mathcal{B}_{r}= \emptyset $.
\WHILE{ $ \mathcal{B}_{c} \neq \emptyset $ }
\STATE $ A =  \textrm{argmin}_{B'\in \mathcal{B}_{c}} |B'| $
\STATE $ \mathcal{B}_{c}:= \mathcal{B}_{c}-\{ A \} $
\FORALL{ $ B \in \mathcal{B}_{c} $ }
\STATE For $ A $ and $ B $, find the smallest enclosing convex bounding box $ C $.
\IF{ $ \left(|A|+|B|\right) \geq  |C| $ }
\STATE $ A = C $
\STATE $ \mathcal{B}_{c} := \mathcal{B}_{c} - \{ B \} $
\ENDIF
\ENDFOR
\STATE $ \mathcal{B}_{r}  := \mathcal{B}_{r} \cup \{ A \} $
\ENDWHILE
\STATE \textbf{return} $ \mathcal{B}_{r} $
\end{algorithmic}
\end{algorithm}

In order to mitigate severe biases and heavy overlaps of foreground sub-areas in coarse detection, we expand the width and height of each detected bounding box from the center with an expansion ratio $\beta$ to roughly enclose its ground truth. Thereafter, we propose a greedy Foreground Region Generation (FRG) algorithm to merge the expanded results as summarized in Algorithm \ref{Cluster Region Generating algorithm}. 

Specifically, FRG takes the expanded coarse detection results $\mathcal{B}_{c} $ as input and selects the box $A$ with the minimal size as the generation starting point. For each remaining box $B$ in $\mathcal{B}_{c} $, FRG searches the smallest convex box $C$ that can enclose $A$ and $B$. 
If the sum of the areas \emph{w.r.t.} $A$ and $B$, \emph{i.e.} $|A|+|B|$, is larger than that of $C$, we update $A$ with $C$ and remove $B$ from $\mathcal{B}_{c}$. This process is repeated until there is no box $B$ that satisfies the condition $|A|+|B|\geq|C|$. In this case, $A$ is collected as a cluster region in $\mathcal{B}_{r}$. We repeat the procedure above until $\mathcal{B}_{c} $ turns to an empty set, and obtain the final merged region set $\mathcal{B}_{r}$.

\subsubsection{Foreground Region Scale Equalization}
After FRG, each image is represented as several cluster regions with different scales. To equalize their scales, especially the small ones, we first estimate the averaged region scale from $\mathcal{B}_{r}$ and subsequently enlarge the regions smaller than a fixed size (\emph{e.g.} $96 \times 96$ in our work) by adjusting the average scale to the fixed one. 


\subsubsection{Foreground Region Packing}
The current approaches individually perform fine-grained detection on each cluster region, which is extremely inefficient. To avoid this, we splice all the regions into a unified mosaic by using the PHSPPOG method \cite{zhang2016priority}. By this mean, fine-grained detection operates only once, remarkably saving the time cost.




\subsection{Multi-Proxy Detection Network}

\begin{figure}[t]
    \centering
    \includegraphics[width=1\columnwidth]{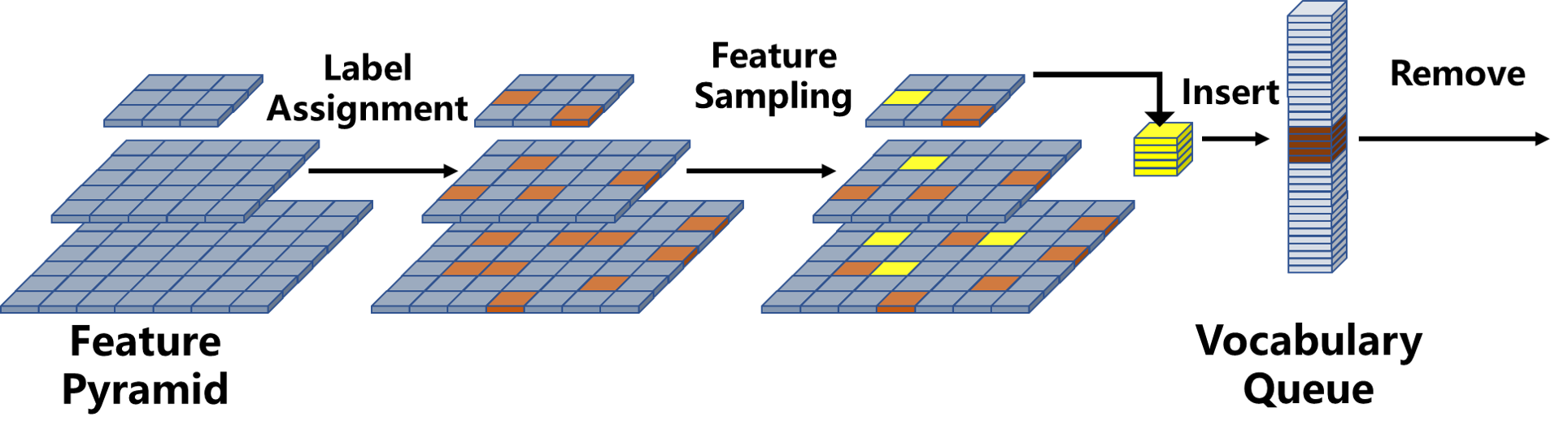}
    \caption{Process of vocabulary queue update. Orange cells indicate positive samples after label assignments, and yellow cells are randomly selected to update the oldest ones in the vocabulary queue.}
    \label{bow update}
    \vspace{-3mm}
\end{figure}

The MP-Det module targets on relieving the confusion between inter-class similarities and intra-class variations from complex object distributions in classification. It includes two main components: (1) the Multi-Proxy Classification Head and (2) the Bag-of-Instance-Words (BoIW) model.


\subsubsection{Multi-Proxy Classification Head}~The conventional classification head assigns a single weight vector $\bm{w}_{i}$ to the $i$-th category. Suppose that $\bm{x}$ is the feature extracted by the backbone network, its conditional probability corresponding to the $i$-th category is formulated as:
\begin{equation}
    Pr(Y=y_{i}|\bm{x}) = \textrm{Sigmoid}(\bm{w}_{i}^{\mathrm{T}}\bm{x}),
    \label{eq:cls}
\end{equation}

\noindent where $\rm{Sigmoid}(\cdot)$ is the sigmoid function. Eq.~\eqref{eq:cls} can be considered as a single proxy classifier, implicitly assuming that each category has only one class center at ${\bm{w}_{i}}/{ \parallel \bm{w}_{i} \parallel }$. 

By contrast, in drone images, intra-class variance is rather large due to great size and view changes of instances. In this case, data belonging to a single class may span around multiple centers. Therefore, we adopt the multi-proxy classification head, by assuming that the $i$-th category has $K$ ($K>1$) proxies denoted by $[\bm{w}_{i}^{1},\bm{w}_{i}^{2},...,\bm{w}_{i}^{K}]$. Inspired by SoftTriple \cite{SoftTriple}, the multi-proxy conditional probability is formulated as:

\begin{equation}
\label{eq:mpd}
\begin{aligned}
     Pr(Y=y_{i}|\bm{x}) & = \textrm{Sigmoid}\left(\gamma \sum_{k} \frac{\exp(s_{i}^{k})s_{i}^{k}}{\sum_{l=1}^{K} \exp(s_{i}^{l})}\right),\\
\end{aligned}
\end{equation}
where $s_{i}^{k}  = \frac{{\bm{w}_{i}^{k}}^{\mathrm{T}}\bm{x}}{\| \bm{w}_{i}^{k}\| \| \bm{x} \| }$ and $\gamma$ is a scaling factor. Accordingly, the weights $[\bm{w}_{i}^{1},\bm{w}_{i}^{2},...,\bm{w}_{i}^{K}]$ can be optimized by minimizing the cross-entropy or the focal loss \emph{w.r.t.} $Pr(Y=y_{i}|x)$.

As in Eq.~\eqref{eq:mpd}, the decision boundary for the $i$-th class is represented by multiple centers (proxies), thus being more flexible and accurate than that by a single one. 

\subsubsection{Bag-of-Instance-Words guided Optimal Transport} The multiple proxies tend to collapse into a similar one during model training, caused by extremely imbalanced cluster assignments. To overcome this dilemma, an optimal transport based procedure is adopted as in \cite{MPWC}. Specifically, given $N_{i}$ positive instance features  $\{\bm{f}_{i}^{j}\}_{j=1}^{N_{i}}$ and $K$ proxies $\{\bm{w}_{i}^{k}\}_{k=1}^{K}$ for the $i$-th class, the cost matrix is computed as $\bm{C} =\frac{\bm{1}- \bm{S}}{2}$, where $\bm{S} \in \mathbb{R}^{N_{i}\times K}$ refers to the cosine similarity matrix between $\{\bm{f}_{i}^{j}\}_{j=1}^{N_{i}}$ and $\{\bm{w}_{i}^{k}\}_{k=1}^{K}$ and $\bm{1}$ is an all-ones matrix. A transportation plan $\bm{P}^{*}_{i} \in \mathbb{R}^{N_{i} \times K}$ is then built as:

\begin{equation}
\label{eq:otp}
\begin{aligned}
 &\bm{P}^{*}_{i} = {\arg\min}_{\bm{P}}  ~~\mathrm{tr}\left(\bm{C}_{i}^{T}\bm{P}\right) \\
 s.t.~~~\sum_{j=1}^{N_{i}}&\bm{P}(j,k)=\bm{p}_{i}(k), \sum_{k=1}^{K}\bm{P}(j,k)=\bm{q}_{i}(j), 
\end{aligned}
\end{equation}

\noindent where $\mathrm{tr}(\cdot)$ is the matrix trace. $\bm{p}_{i}=[\bm{p}_{i}(1), \cdots, \bm{p}_{i}(K)]$ and $\bm{q}_{i}=[\bm{q}_{i}(1), \cdots, \bm{q}_{i}({N_{i}})]$ indicate the marginal probability distributions of $\{\bm{w}_{i}^{k}\}_{k=1}^{K}$ and $\{\bm{f}_{i}^{j}\}_{j=1}^{N_{i}}$, respectively. Based on Eq.~\eqref{eq:otp}, the instance-proxy matching loss \emph{w.r.t.} $\{\bm{f}_{i}^{j}\}_{j=1}^{N_{i}}$ and $\{\bm{w}_{i}^{k}\}_{k=1}^{K}$ is defined as

\begin{equation}
\label{eq:loss_ot}
 \mathcal{L}_{ot} = \frac{1}{N_{c}}\sum_{i=1}^{N_{c}} \mathrm{tr}\left(\bm{C}_{i}^{T}\bm{P}_{i}^{*}\right),
\end{equation}

\noindent where $N_{c}$ is the number of classes. 




\begin{table*}[!t]
	\centering
	\begin{center}
		\begin{tabular}{c|c|ccc|ccc|ccc}
			\hline
			\multirow{2}{*}{Method} & \multirow{2}{*}{References} &  \multicolumn{3}{c|}{\underline{ResNet-50}} & \multicolumn{3}{c|}{\underline{ResNet-101}} &
			\multicolumn{3}{c}{\underline{ResNeXt-101}}\\ 
			&  & AP & AP50 & AP75 & AP & AP50 & AP75 & AP & AP50 & AP75 \\ 
			\hline
			Faster-RCNN & \cite{faster-rcnn} & 21.4 & 40.7 & 19.9 & 21.4 & 40.7 & 20.3 & 21.8 & 41.8  & 20.1 \\
			ClusDet & \cite{ClutDet}  & 26.7  & 50.6  & 24.4 & 26.7 & 50.4 & 25.2 & 32.4 & 56.2 & 31.6 \\
			DMNet & \cite{DMNet} & 28.2  & 47.6  & 28.9 & 28.5 & 48.1 & 29.4 & 29.4 & 49.3 & 30.6 \\
			GLSAN & \cite{GLSAN} & 30.7  & 55.4  & 30.0 & 30.7 & 55.6 &  29.9 & - & - & - \\
			SAIC-FPN & \cite{SAIC-FPN}  & -  & -  & - & - & - &  - & 35.7 & 62.3 & 35.1 \\
			AMRNet & \cite{ARMNet}  & 31.7  & 52.7  & 33.1 & 31.7 & 52.6 &  33.0 & 32.1 & 53.0 & 33.2 \\
			HRDNet & \cite{HRDNet}  & -  & -  & - & 31.4 & 53.3 &  31.6 & 35.5 & 62.0 & 35.1 \\
			\hline
			\textbf{UFPMP-Det} & Ours & 36.6 & 62.4 & 36.7 & 37.5 & 63.2  & 38.3 & 39.2 & 65.3 & 40.2 \\
			\textbf{UFPMP-Det}+MS & Ours & \textbf{37.4} & \textbf{63.7} & \textbf{37.7} & \textbf{38.7} & \textbf{65.1} & \textbf{39.4} & \textbf{40.1} & \textbf{66.8}  & \textbf{41.3} \\
			\hline
		\end{tabular}
		\caption{Comparison of different approaches in AP/AP50/AP75 (\%) on the validation set of VisDrone. MS refers to the multi-scale trick during inference and `-’ indicates that the result is not reported.}
		\label{table:visdrone_result}
	\end{center}
\end{table*}


In Eq.~\eqref{eq:otp}, the uniform distribution, \emph{i.e.} $\bm{p}_{i}(k)=\frac{1}{K}$ and $\bm{q}_{i}(j)=\frac{1}{N_{i}}$, is usually used as a prior, but this may be different from the case of real data, thus leading to performance drop. To alleviate the misalignment, we develop the BoIW model to estimate the intra-class distribution of $\{\bm{w}_{i}^{k}\}_{k=1}^{K}$. 

Concretely, BoIW firstly constructs a queue-based vocabulary $\bm{V}_{i}$ of size $N$, as the representative feature set for the $i$-th class. As shown in Fig. \ref{bow update}, $\bm{V}_{i}$ is updated in each mini-batch, where $m$ positive instances are selected, and their features are inserted into $\bm{V}_{i}$ by removing the $m$ oldest ones at the mean time. For a two-stage detector, we extract the flattened instance-level features after the RoI layer. As to the one-stage detector, we additionally employ a convolutional layer to extract $C$-dimensional instance features.

Subsequently, $K$-means is applied to $\bm{V}_{i}$ and $K$ clusters $\{\bm{c}_{i}^{k}\}$ are obtained. The marginal distribution $\bm{p}_{i}$ is therefore estimated as $\bm{p}_{i}(k)=\#|\bm{c}_{i}^{k}|/\#|\bm{V}_{i}|$, where $\#|\cdot|$ denotes the number of elements. Considering that the clusters $\{\bm{c}_{i}^{k}\}$ vary in different steps, we sort $\bm{p}_{i}$ in the descending order to ensure that $\bm{w}_{i}^{k}$ always corresponds to the cluster with the $k$-th highest probability. By taking $\bm{p}_{i}(k)$ back to Eqs.~\eqref{eq:otp} and \eqref{eq:loss_ot}, we have the BoIW induced optimal transport loss. 

Note that the vocabulary $\bm{V}_{i}$ is the representative feature set for the $i$-th class. To further enhance their correlations, we adopt another contrastive learning loss \emph{w.r.t.} $\bm{V}_{i}$ and the positive instance feature $\bm{x}_{i}$:
\begin{equation}
\label{eq:cl}
    \mathcal{L}_{cl} = - \frac{1}{N} \sum_{i}\log \left(\frac{\sum_{\bm{v}\in \bm{V}_{i}}\exp(\bm{v}^{\mathrm{T}}\bm{x}_{i})}{\sum_{\bm{u} \in \bm{V}}\exp(\bm{u}^{\mathrm{T}}\bm{x}_{i}) }\right),
\end{equation}
where $\bm{V}=\cup_{i=1}^{N_{c}} \bm{V}_{i}$. 

By minimizing $\mathcal{L}_{cl}$ in Eq.~\eqref{eq:cl}, the correlations between the intra-class/inter-class features are increased/decreased.


\subsubsection{Adaptive $\bm{K}$-Proxy Estimation}
The number of proxies $K$ within each class is important to MP-Det. Instead of the trivial solution to set it manually, we propose an adaptive way to estimate $K$. To be specific, we extract instance features by the vanilla GFL v1 \cite{GFLV1} and perform clustering by DBSCAN \cite{dbscan} to determine $K$.

\subsubsection{Overall Optimization}
Finally, the overall training loss of MP-Det combines the conventional detection loss $\mathcal{L}_{det}$, the BoIW induced optimal transportation loss $\mathcal{L}_{ot}$ in Eq.~\eqref{eq:otp} and the contrastive learning loss $\mathcal{L}_{cl}$ in Eq~\eqref{eq:cl}, formulated as:
\begin{equation}
    \mathcal{L} = \mathcal{L}_{det} + \mathcal{L}_{ot}+\mathcal{L}_{cl}.
\end{equation}

During training, we perform BoIW to regularly estimate $\bm{p}_{i}$ (for every 2,000 iterations in our case). When optimizing $\mathcal{L}_{ot}$, we employ Sinkhorn-Knopp \cite{Cuturi13} to compute the transportation plan $\bm{P}^*_{i}$, which only slightly increases the training time (27.5h \emph{vs.} 33h) without any extra cost at test.

\vspace{-3mm}
\section{Experimental Results and Analysis}

UFPMP-Det is evaluated on the widely-used \textbf{VisDrone} \cite{VisDrone} and \textbf{UAVDT} \cite{UAVDT} benchmarks and extensive experiments are carried out.

\subsection{Datasets and Protocols}
\textbf{VisDrone} consists of 10,209 high resolution images ($2000 \times 1500$) with 10 object categories, captured by various drone-mounted cameras from different areas (\emph{urban} and \emph{country}) and scenes (\emph{sparse} and \emph{crowded}). 6,471 images are used for training, 548 for validation and 3,190 for test. Since the test set is not publicly available, we follow ClutDet \cite{ClutDet} and DMNet \cite{DMNet} to report scores on the validation set. \textbf{UAVDT} includes 23,258 images for training and 15,069 images for test. All the images are captured from urban areas by a UAV at low altitudes with a $1080\times 540$ resolution. Three kinds of vehicles (\emph{car}, \emph{bus}, and \emph{truck}) are manually labeled. Similar to the protocols for general object detection \cite{COCO}, we adopt Average Precision (AP) and APs at the IoU threshold of 0.5 (AP50) and 0.75 (AP75) as the metrics on both the datasets. 


\begin{table}[!t]
\centering
\resizebox{1\linewidth}{!}{
\begin{tabular}{c|c|ccc}
\hline
Method  & Reference  & AP & AP50 & AP75
\\ 
\hline
Faster-RCNN   & \cite{faster-rcnn}  & 11.0    & 23.4 & 8.4  \\
ClusDet   & \cite{ClutDet}  & 13.7    & 26.5 & 12.5  \\
DMNet   & \cite{DMNet}  & 14.7    & 24.6 & 16.3  \\ 
GLSAN   & \cite{GLSAN}  & 17.0    & 28.1 & 18.8  \\
DREN   & \cite{DREN}  & 15.1    & - & -  \\
ARMNet   & \cite{ARMNet}  & 18.2    & 30.4 & 19.8  \\\hline
\textbf{UFPMP-Det}   & Ours  & \textbf{24.6}    & \textbf{38.7} & \textbf{28.0}  \\ \hline 
\end{tabular}
}
\caption{Comparison of different approaches with ResNet-50 in AP/AP50/AP75 (\%) on UAVDT. `-' indicates that the result is not reported.}
\label{table:uavdt_result}
\end{table}


\subsection{Implementation Details}
We implement the proposed approach using the open-source \emph{MMDetection} toolbox$\footnote{https://github.com/open-mmlab/mmdetection}$. GFL \cite{GFLV1} is employed as the baseline detector with the model pre-trained on ImageNet. UFPMP-Det is trained for 60 epochs in total by the SGD optimizer. The momentum and weight decay are fixed as 0.9 and 0.0001, respectively. The initial learning rate is set at 0.01 with a linear warm-up, which decreases by the factor of 10 after 40 and 55 epochs. Regarding BoIW model learning, it is individually updated in the first 10 epochs without performing optimal transport and contrastive learning, and jointly optimized for all the components afterwords, which is empirically stable during training in our experiments. The size of the input image in our detector is set to $1333 \times 800$ on VisDrone and $1000 \times 600$ on UAVDT.



\begin{table}[!t]
\centering
\resizebox{0.98\linewidth}{!}{
\begin{tabular}{c|c|c|c}
\hline
Method  & Reference & \#img & Inference Time \\
\hline
ClusDet   & \cite{ClutDet} & 2716 & 0.273  \\
DMNet & \cite{DMNet} & 2736 & 0.290 \\
\hline
\textbf{UFPMP-Det} & Ours & \textbf{1096}& \textbf{0.152}  \\
\hline
\end{tabular}
}
\caption{Comparison of different methods in efficiency \emph{w.r.t.} the number of packed images (\#img) and the inference time cost (in seconds) on VisDrone.}
\label{tab:efficiency}
\end{table}

\subsection{Comparison with the State-of-the-arts}
We compare UFPMP-Det with some state-of-the-art counterparts, including Faster-RCNN \cite{faster-rcnn}, ClusDet \cite{ClutDet}, DMNet \cite{DMNet}, GLSAN \cite{GLSAN}, and DREN \cite{DREN}.

\noindent \textbf{Results on VisDrone.} Existing approaches deliver results by using different base networks on VisDrone, and we therefore report the performance of UFPMP-Det with various typical backbones, \emph{i.e.} ResNet-50, ResNet-101, and ResNeXt-101, for more comprehensive validation. As summarized in Table \ref{table:visdrone_result}, all these methods generally achieve higher accuracies through stronger networks. When using the same backbone, UFPMP-Det consistently outperforms such counterparts by large margins, improving APs of the second best by 4.9\%, 5.8\% and 3.5\% with ResNet-50, ResNet-101 and ResNeXt-101, respectively. It is worth noting that the performance of UFPMP-Det with ResNet-50 is even superior to that with much deeper networks (\emph{e.g} ResNeXt-101), reaching the new state-of-the-art. Besides, the multi-scale technique during inference further promotes the accuracy.

\noindent \textbf{Results on UAVDT.} Most of detectors utilize the ResNet-50 backbone for evaluation on UAVDT, and we follow this setting for fair comparison. As illustrated in Table \ref{table:uavdt_result}, UFPMP-Det largely boosts the performance of other detectors, and it improves the AP, AP50 and AP75 of the second best ARMNet by 6.4\%, 8.3\% and 8.2\%, respectively. 


\noindent \textbf{Overall Complexity.} To analyze the efficiency of UFPMP-Det, we show the number of packed images as well as the inference time cost, in comparison to ClusDet \cite{ClutDet} and DMNet \cite{DMNet}. All the experiments are conducted on one GTX 1080TI GPU. As Table \ref{tab:efficiency} displays, UFPMP-Det generates less than half of the packed images by ClusDet and DMNet and thus infers significantly faster, highlighting its advantage.


\subsection{Ablation Study}
We detailedly validate the major components, \emph{i.e.}, UFP and MP-Det, as well as several hyper-parameter on UFPMP-Det.

\begin{table}[!t]
\begin{center}
\resizebox{\linewidth}{!}{
\begin{tabular}{c|c|c|ccc}
\hline
Method  & Reference & \#img & AP & AP50 & AP75    \\ \hline
EIP   & \cite{ClutDet} & 3288 & 21.1    & 44.0 & 18.1  \\
ClusDet   & \cite{ClutDet} & 2716 & 26.7    & 50.6 & 24.7  \\
DMNet & \cite{DMNet} & 2736 & 28.2    & 47.6 & 28.9  \\
\hline 
\textbf{UFP} & Ours & \textbf{1096} & \textbf{30.6} & \textbf{52.5} & \textbf{31.0} \\
\hline
\end{tabular}
}
\end{center}
\caption{Comparison of packing methods in AP/AP50/AP75 (\%) and number of packed images (\#img) based on Faster-RCNN with ResNet-50.}
\label{ufp overall}
\end{table}

\begin{table}[!t]
\begin{center}
\resizebox{0.98\linewidth}{!}{
\begin{tabular}{c|c|c|ccc}
\hline
Dataset  & UFP & FR & Small & Medium & Large \\ \hline
VisDrone &  & 10.2 & 68.56 & 28.68 & 2.76 \\
VisDrone & \checkmark & 24.5 & 6.96 & 63.35 & 29.69 \\ \hline
UAVDT &  & 5.11 & 74.87 & 23.01 & 2.12 \\
UAVDT & \checkmark & 22.98 & 0.64 & 71.60 & 27.76 \\
\hline
\end{tabular}
}
\end{center}
\caption{Ablation study on UPF \emph{w.r.t.} Foreground Ratio (FR) (\%) and proportion (\%) of object instances in small, medium and large sizes (using the COCO metric) on VisDrone and UAVDT.}
\label{statistic ufp}
\vspace{-3mm}
\end{table}

\begin{figure}[!t]
\centering

\subfigure{\includegraphics[width=4.0cm]{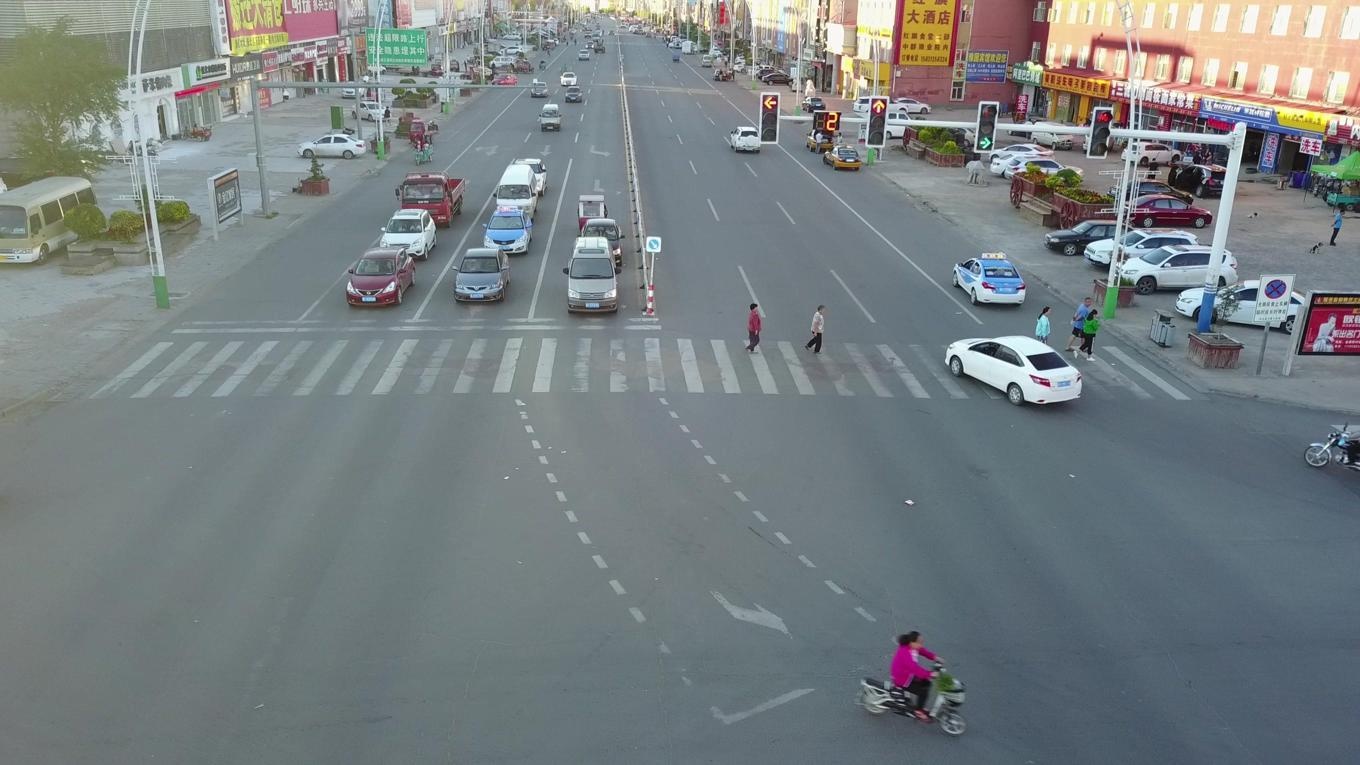}} 
\subfigure{\includegraphics[width=4.0cm]{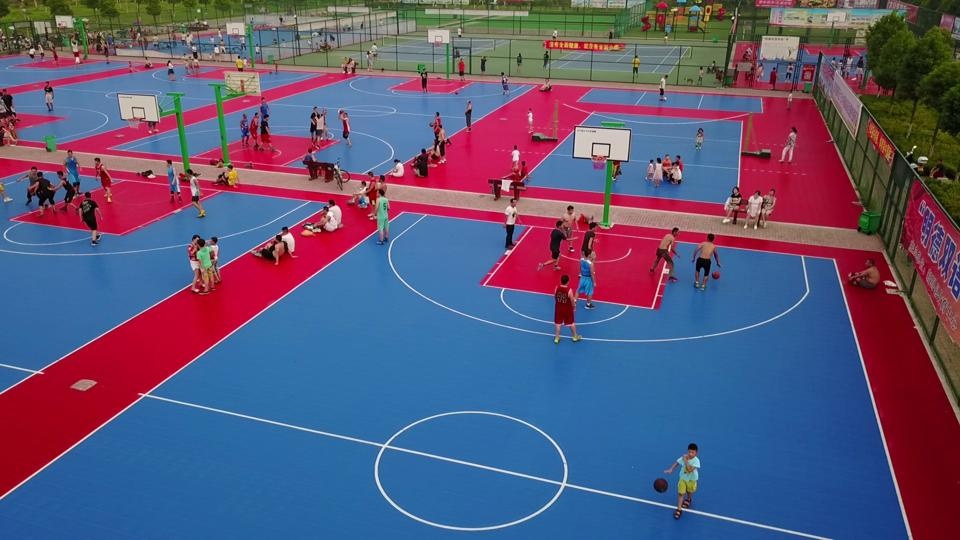}}
\\ 
\vspace{-3mm}
\centering
\subfigure{\includegraphics[width=4.0cm]{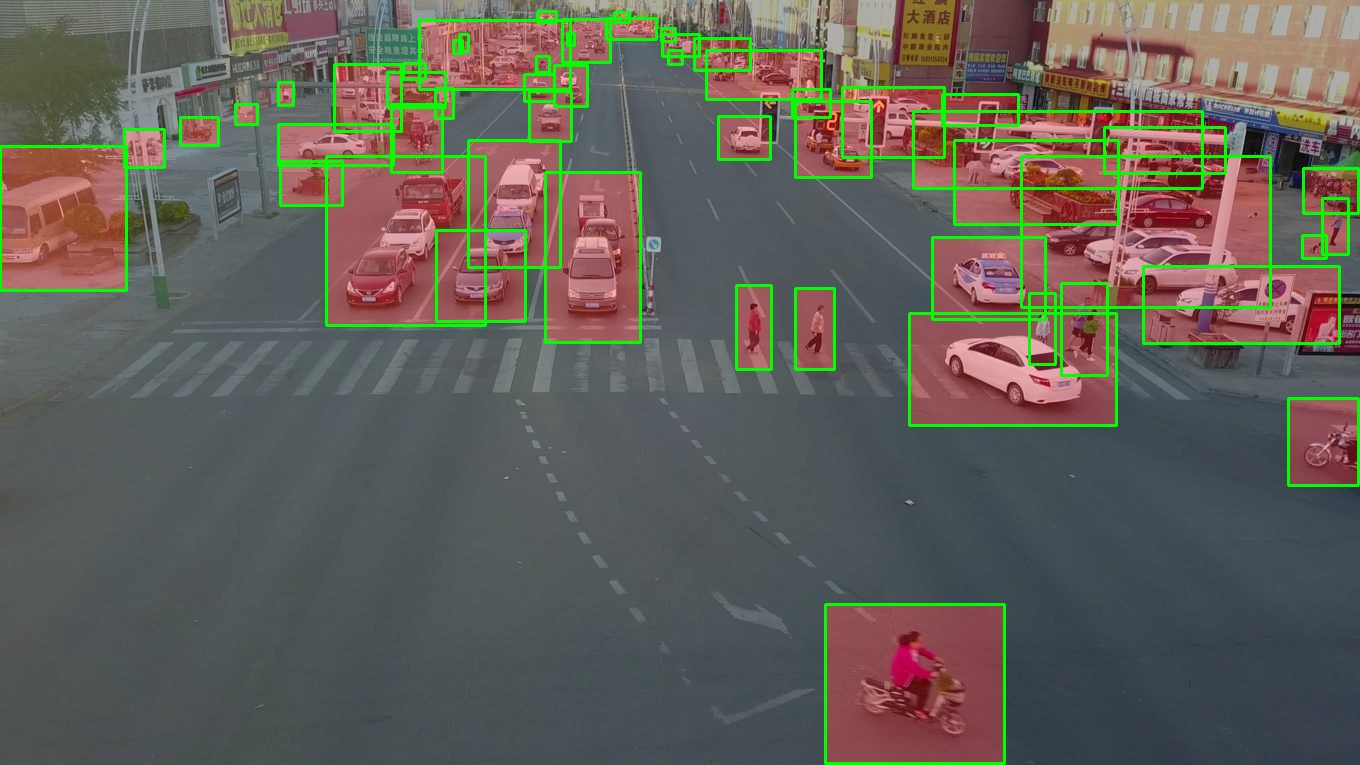}} 
\subfigure{\includegraphics[width=4.0cm]{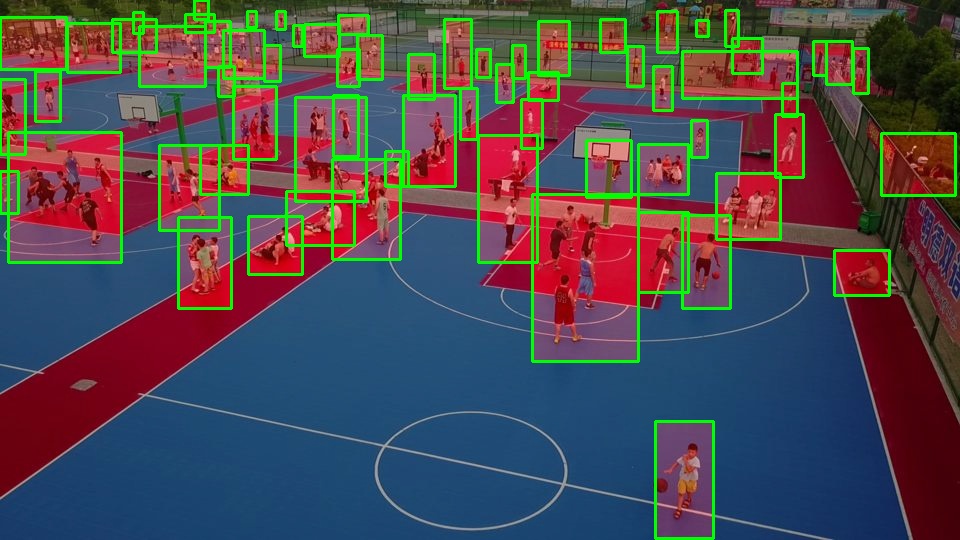}}
\\
\vspace{-3mm}
\centering
\subfigure{\includegraphics[width=4.0cm, height=4.0cm]{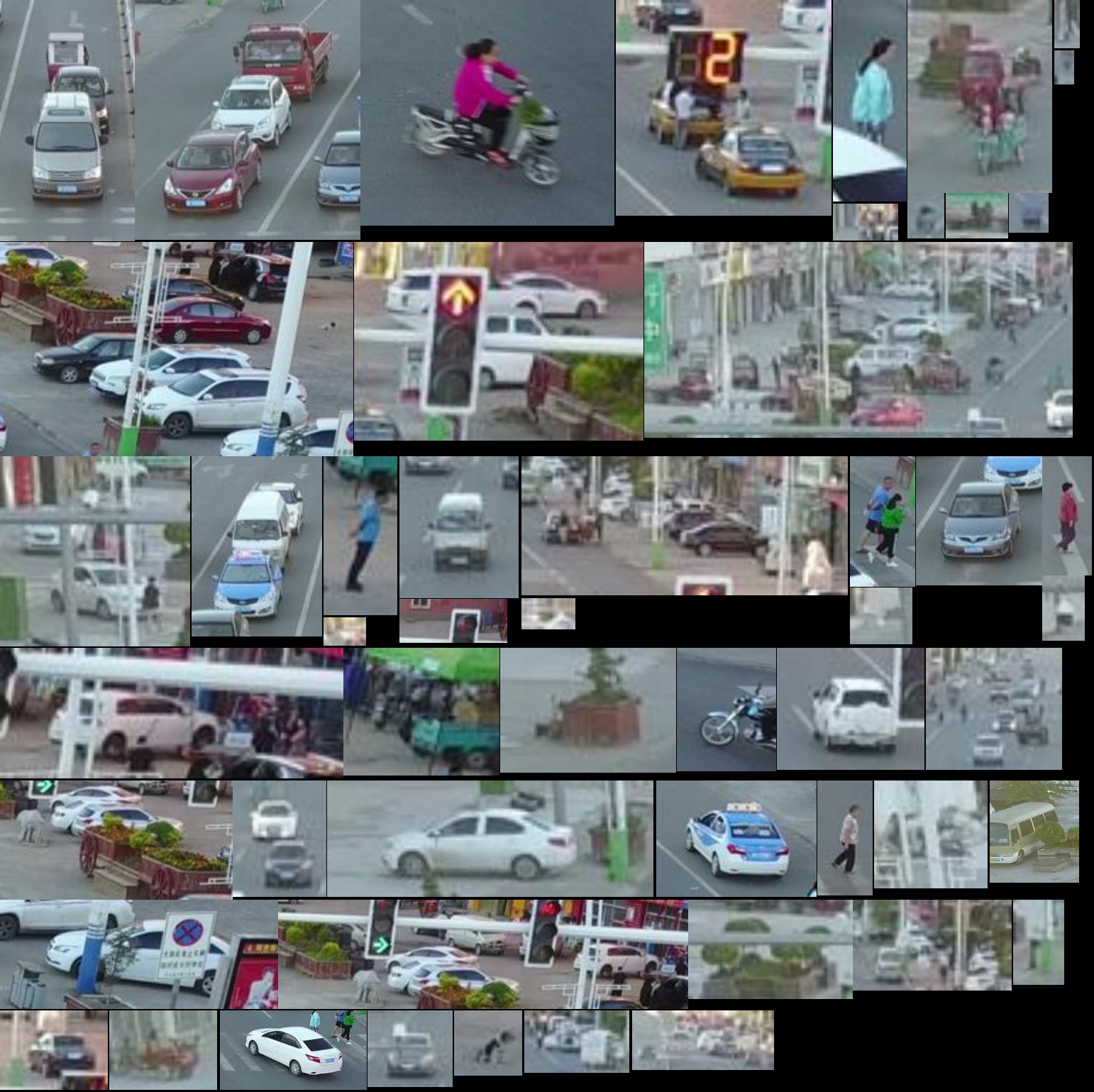}} 
\subfigure{\includegraphics[width=4.0cm, height=4.0cm]{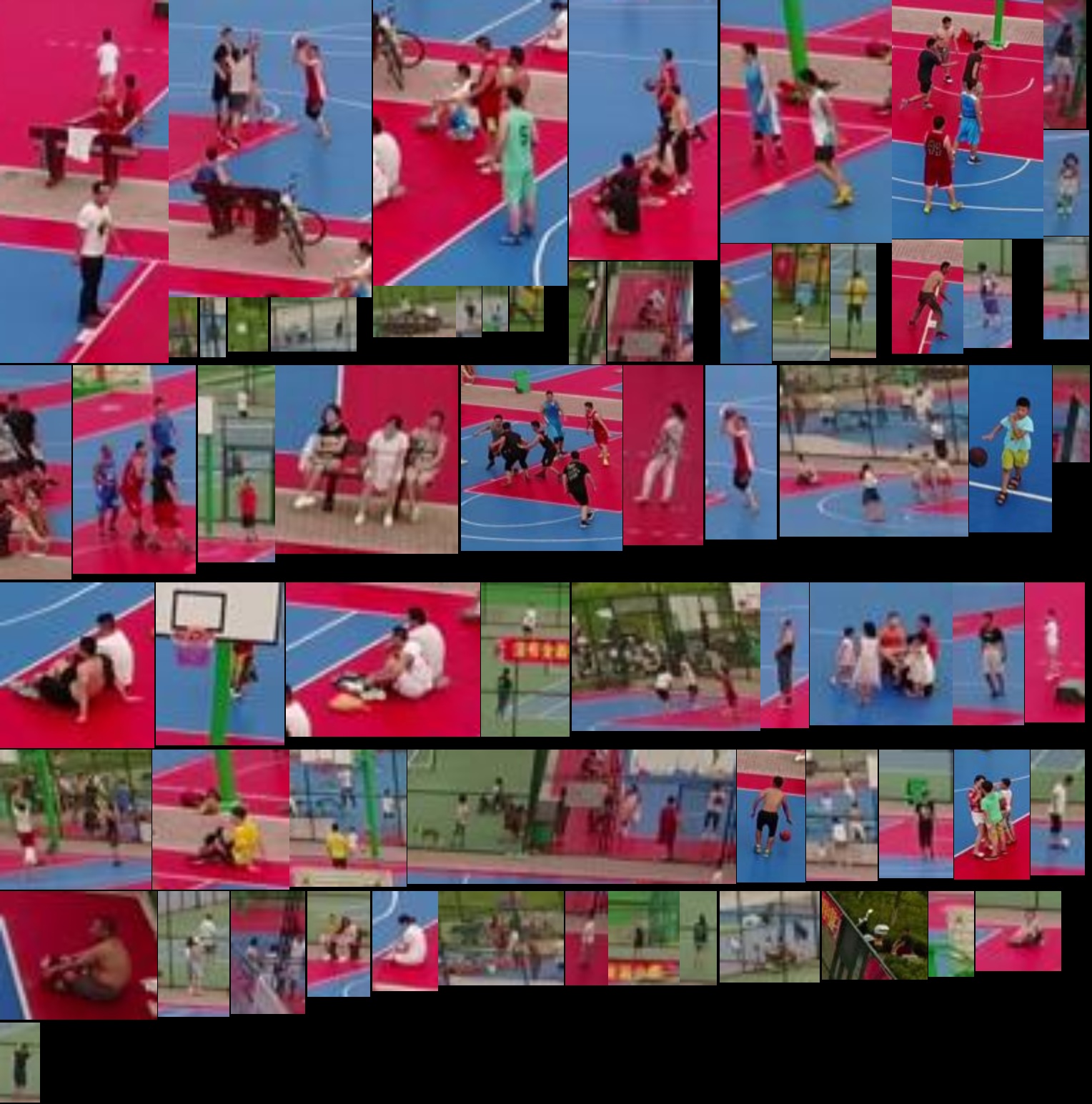}}
\caption{Visualization of UFP. Top: input images; middle: clustered regions highlighted by green bounding boxes; and bottom: packed mosaics.} 
\vspace{-3mm}
\end{figure}
\subsubsection{On UFP.} We evaluate UFP and compare it to three counterparts: \emph{i.e.} evenly image partition (EIP) \cite{ClutDet}, ClusDet \cite{ClutDet}, and DMNet \cite{DMNet}. For fair comparison, all the methods utilize Faster-RCNN with FPN as the base detector and ResNet-50 as the backbone. As shown in Table \ref{ufp overall}, UFP outputs less packed images by performing unified adaptive packing, whilst achieving the best accuracy by increasing the Foreground Ratio (FR). To show the advantage of UPF in increasing FR and decreasing the number of small objects, we summarize the FR as well as the percentages of small/medium/large objects on VisDrone and UAVDT in Table \ref{statistic ufp}. It is worth noting that we utilize the metric used in MS COCO to determine whether an object is small, medium or large. As demonstrated, UFP clearly promotes FR and significantly reduces the percentage of small objects, thereby facilitating successive detection. 

We visualize the intermediate and the final outputs of UFP in Fig. \ref{ufp overall}. The first row shows the input images; the second row depicts the FRG clusters from the object regions densely extracted by the foreground detector; and the last row displays the packed unified mosaics.


\begin{table}[!t]
\begin{center}
\resizebox{1.0\linewidth}{!}{
\begin{tabular}{l|ccc}
\hline
Method  & AP & AP50 & AP75 \\ \hline 
Baseline  & 29.6  & 49.8  & 30.3 \\ 
Baseline+UFP & 36.6  & 62.3  & 36.8 \\ 
Baseline+UFP+MP-Head & 37.0  & 62.5  & 37.6 \\ %
Baseline+UFP+MP-Head+BoIW  & 37.5  & 63.2  & 38.3 \\ 
\hline

\end{tabular}
}
\end{center}
\caption{Validation of different components in MP-Det with ResNet-101 on VisDrone in terms of AP/AP50/AP75 (\%).}
\label{mp-det}
\end{table}

\subsubsection{On MP-Det.} MP-Det has two major components: \emph{i.e.} MP-Head and BoIW, and we validate them on a stronger baseline with ResNet-101. As shown in Table \ref{mp-det}, MP-Head first improves the baseline accuracy (AP) by 0.4\% and BoIW further increases it by 0.5\%, highlighting their credits.



\begin{table}[!t]
\begin{center}
\begin{tabular}{c|ccc}
\hline
Method         &  AP  & AP50 & AP75 \\ \hline
$K=10$ (Manual) & 37.1 & 62.8 & 37.8 \\
$K=20$ (Manual) & 37.1 & 62.6 & 37.7 \\
\hline
MP-Det & 37.5 & 63.2 & 38.3 \\
\hline
\end{tabular}
\end{center}
\caption{Comparison of AP/AP50/AP75 (\%) on VisDrone by using different methods to set the number of proxies $K$: Manual \emph{vs.} MP-Det (Ours).}
\label{k-proxies}
\end{table}

\begin{table}[!t]
\begin{center}
\begin{tabular}{c|cccc}
\hline
Size of BoIW ($N$)   & AP & AP50 & AP75 \\ \hline
50 & 22.3 & 36.1 & 25.2 \\
100 & 24.0& 37.6 & 27.1  \\
200 & 24.6 & 38.7 & 28.0 \\
\hline
\end{tabular}
\end{center}
\caption{The impact of the size of BoIW ($N$) on the performance of MP-Det \emph{w.r.t.} AP/AP50/AP75 (\%) on UAVDT.}
\label{size of boiw}
\end{table}

We further evaluate the effects of adaptive $K-$proxy estimation and the size of BoIW on the performance of MP-Det. As aforementioned, MP-Det adaptively estimates the number of proxies for each class by performing DBSCAN on the instance features extracted from pre-trained models. For comparison, we choose the way to set it manually as the baseline. As shown in Table \ref{k-proxies}, manually setting is not as good as adaptively setting when used in MP-Det.  

As to the size of BoIW, we report the results of MP-Det by using different values, \emph{e.g.}, $N=50/100/200$ on UAVDT in Table \ref{size of boiw}. As summarized, MP-Det achieves the highest score with $N=200$, which is therefore adopted in our work. 

\begin{figure}[!t]
\centering
\subfigure[with OT]{\includegraphics[width=0.4\textwidth]{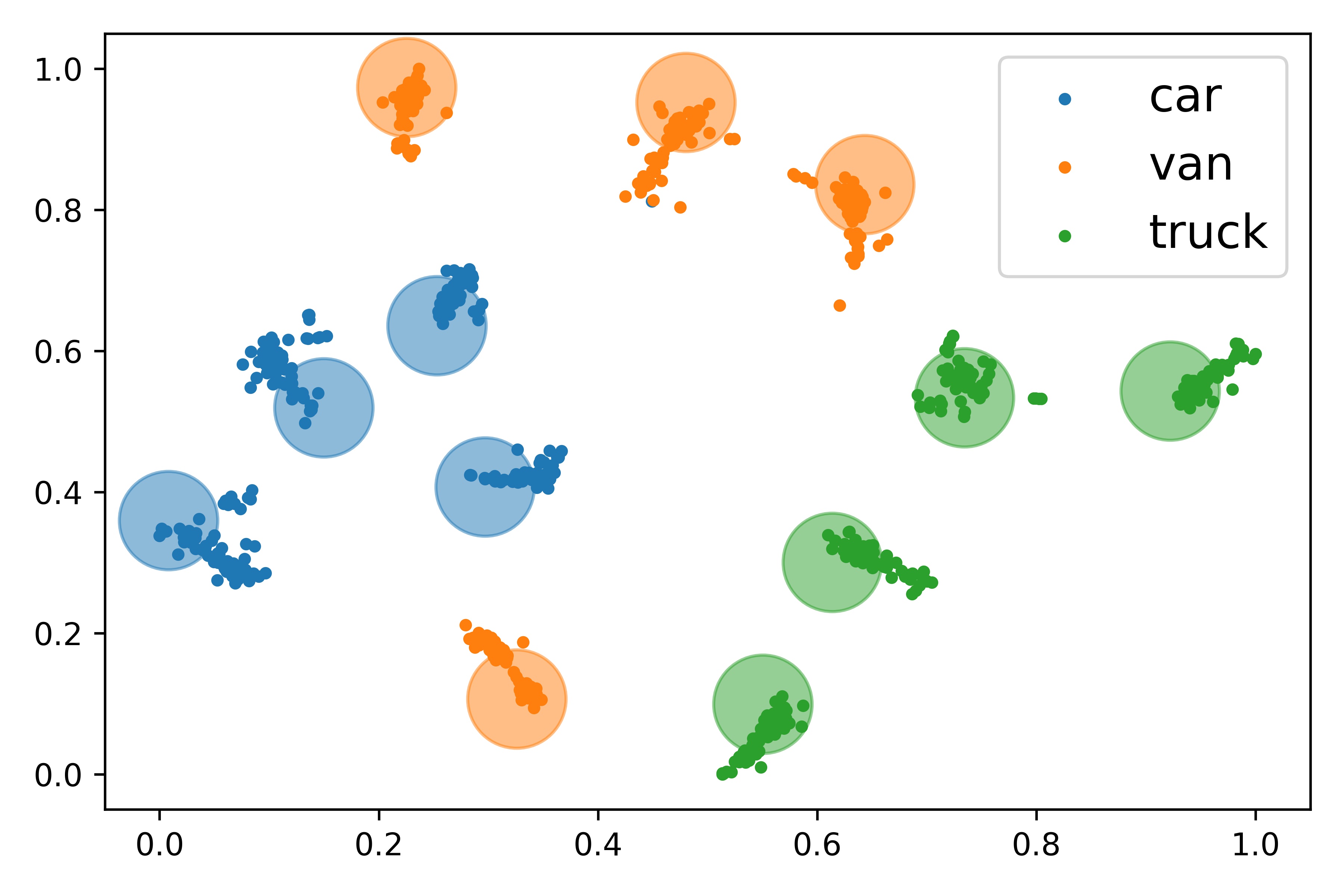}}
\vspace{-3mm}
\subfigure[w/o OT]{\includegraphics[width=0.4\textwidth]{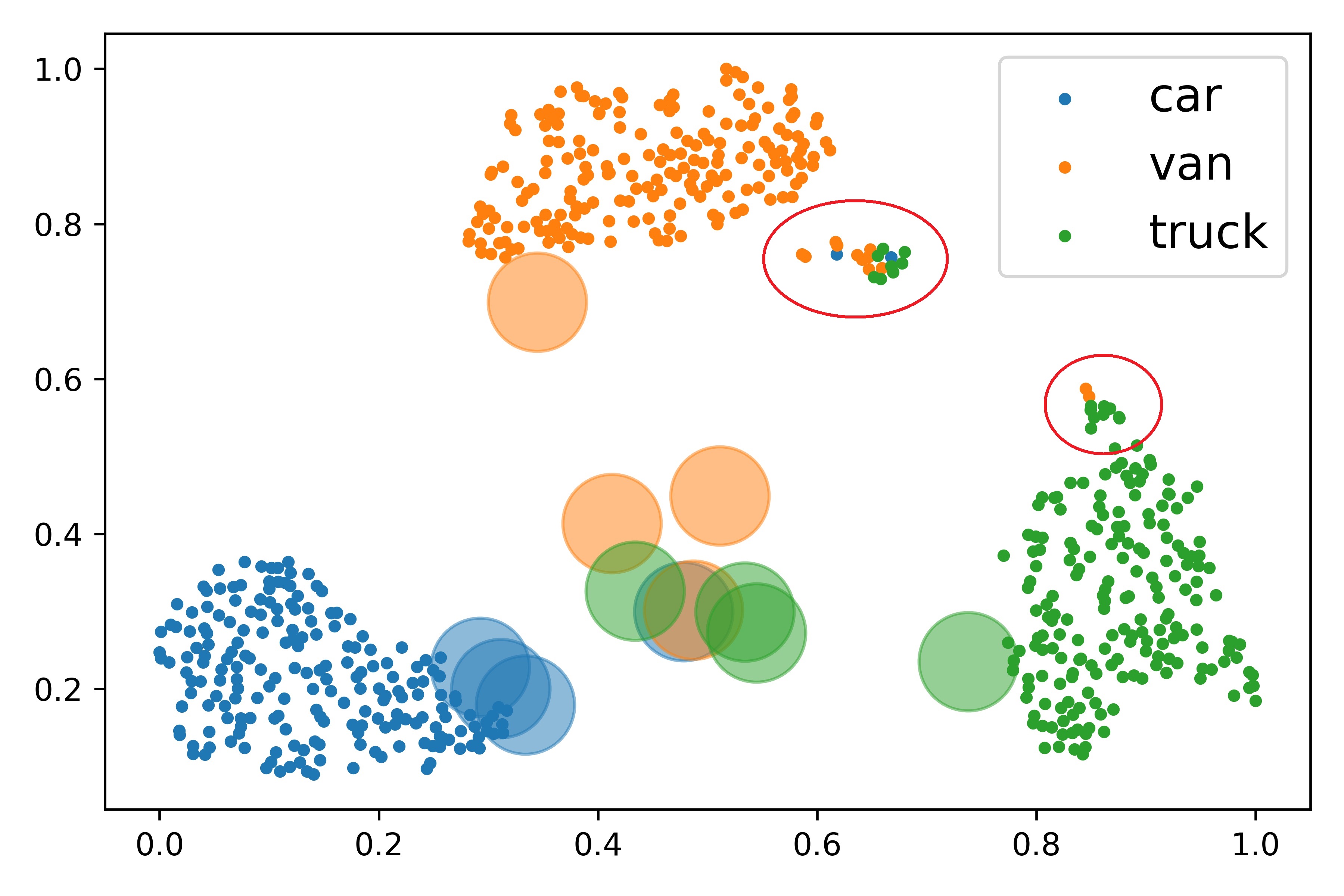}}
\caption{Visualization of multi-proxies learned on UAVDT via $t$-SNE. Solid dots denote the features of object instances and translucent circles denote the proxies. Different colors indicate distinct classes} 
\label{visproxy}
\vspace{-3mm}
\end{figure}

We qualitatively demonstrate the impact of optimal transport (OT) based feature-proxy matching, by visualizing the features of object instances as well as the proxies on UAVDT via $t$-SNE. As shown in Fig.~\ref{visproxy}, without OT, the features of instances tend to gather to non-proxy points and are apart from the proxies, and the proxies from different classes are not discriminative enough. In contrast, when OT is employed, the learned features locate in the small neighborhoods of multiple proxies in a more uniformly distributed way. Besides, the inter-class distances between proxies are clearly enlarged, making it easier for object classification.

\begin{table}[!t]
\begin{center}
\begin{tabular}{c|c|ccc}
\hline
$\beta$ & FR  & AP & AP50 & AP75 \\ \hline
1.3 & 32.46 & 33.9 & 61.7 & 33.5  \\
1.5 & 24.53 & 36.6 & 62.4 & 36.7 \\
1.7 & 21.07 & 35.6 & 61.3 & 35.8 \\
\hline
\end{tabular}
\end{center}
\caption{The impact of $\beta$ on UFP \emph{w.r.t.} Foreground Ratio (FR) and AP/AP50/AP75 (\%) on VisDrone.}
\label{exp beta}
\vspace{-3mm}
\end{table}

\subsubsection{On Hyper-parameter $\bm{\beta}$.} As described, $\beta$ affects the number of clustered regions and the average recall of raw images. In Table \ref{exp beta}, we report the detection accuracies for different values of $\beta$. The results indicate that FR decreases as $\beta$ increases, and UFP reaches the highest performance when $\beta=1.5$, which is therefore used in our experiments. 


\vspace{-3mm}
\section{Conclusion}
In this paper, we propose a novel approach, namely UFPMP-Det, to object detection on drone imagery. It first introduces the UFP module to address the instances of very small scales by generating single mosaics of input images with largely increased foreground ratios, substantially improving both the accuracy and efficiency. The MP-Det module is further presented to model complex object distributions through multiple proxy learning, where the proxies are enforced to be diverse by minimizing a Bag-of-Instance-Words guided optimal transport loss. Extensive experiments are conducted on two benchmarks, and UFPMP-Det reaches the new state-of-the-art, highlighting its effectiveness.

\vspace{-3mm}
\section{Acknowledgment}
This work is partly supported by the National Natural Science Foundation of China (No. 62022011), the Research Program of State Key Laboratory of Software Development Environment (SKLSDE-2021ZX-04), and the Fundamental Research Funds for the Central Universities.
\bibliography{references}

\end{document}


\maketitle

\section{Influence of Base Detector}
\begin{table}[!t]
\centering
\resizebox{1\linewidth}{!}{
\begin{tabular}{c|c|ccc}
\hline
Method  & BaseDetector  & AP & AP50 & AP75
\\ 
\hline
Faster-RCNN   & -   & 11.0    & 23.4 & 8.4  \\
GFL   & -   & 16.4    & 28.3 & 17.4  \\
ClusDet   & FasterRCNN   & 13.7    & 26.5 & 12.5  \\
DMNet   & FasterRCNN   & 14.7    & 24.6 & 16.3  \\ \hline 
\textbf{UFPMP-Det}   & FasterRCNN  & \textbf{20.4}    & \textbf{33.4} & \textbf{22.8}  \\ 
\textbf{UFPMP-Det}   & GFL  & \textbf{24.6}    & \textbf{38.7} & \textbf{28.0}  \\ \hline 
\end{tabular}
}
\caption{Comparison of AP (\%), AP50 (\%) and AP75 (\%) by using various approaches on UAVDT with ResNet-50.} \label{table:uavdt_result}
\end{table}
Since UFP and MP structure are flexible plugins with any base detector, we implement our method on FasterRCNN and GFL, respectively and keep the hyperparameter of the base detector as MMDetection default settings. Existing work utilizes different input size where training and inference. The input size of clusDet is set to $ 1000 \times 1000 $, where DMNet is $ 600 \times 1000 $. Therefore, we utilize the minimal input size to carry out our experiment. 

\textbf{UAVDT} we summary the results in Table. \ref{table:uavdt_result}. Compared with the performance of two base detectors. GFL achieve 16.4\% AP, which outperforms the FasterRCNN than 5.4\% AP. That is because that the quantity-IoU branch eliminates the inconsistency risk and accurately depicts the flexible distribution in the drone image object. However, our method can improve both detectors by a large margin. On FasterRCNN, UFPMP-Det achieves 20.4\% AP, which outperforms baseline by 9.4\%. On GFL, UFPMP-Det can outperform baseline by 8.2\% AP. That demonstrates that our method can flexibly be utilized in the different base detectors and boost performance significantly.